%% file: main.tex
\documentclass{article} 
\usepackage{iclr2025_conference,times}

\input{math_commands.tex}

\usepackage{hyperref}
\usepackage{url}
\usepackage{booktabs}
\usepackage{graphicx}
\usepackage{pgfplots}
\pgfplotsset{compat=1.14}

\title{Gaussian Mixture Vector Quantization with Aggregated 
Categorical Posterior}

\author{
  Mingyuan Yan\thanks{Equal contributions.} \\
  National University of Singapore \\
  Singapore\\
  \texttt{yan167951620@gmail.com} \\
   \And
   Jiawei Wu\footnotemark[1] \\
  National University of Singapore \\
  Singapore\\
  \texttt{jiawei.wu1023@gmail.com} \\
   \And
  Rushi Shah\\
  Indian Institute of Technology Jodhpur \\
  India\\
  \texttt{shah.15@iitj.ac.in} \\
     \And
  Dianbo Liu\thanks{Corresponding author.}\\
   National University of Singapore \\
  Singapore\\
  \texttt{dianbo@nus.edu.sg} \\
}

\begin{document}

\maketitle

\begin{abstract}

The vector quantization is a widely used method to map continuous representation to discrete space and has important application in tokenization for generative mode, bottlenecking information and many other tasks in machine learning. 
Vector Quantized Variational Autoencoder (VQ-VAE) is a type of variational autoencoder using discrete embedding as latent. We generalize the technique further, enriching the probabilistic framework with a Gaussian mixture as the underlying generative model. This framework leverages a codebook of latent means and adaptive variances to capture complex data distributions. This principled framework avoids various heuristics and strong assumptions that are needed with the VQ-VAE to address training instability and to improve codebook utilization.  This approach integrates the benefits of both discrete and continuous representations within a variational Bayesian framework. Furthermore, by introducing the \textit{Aggregated Categorical Posterior Evidence Lower Bound} (ALBO), we offer a principled alternative optimization objective that aligns variational distributions with the generative model. Our experiments demonstrate that GM-VQ improves codebook utilization and reduces information loss without relying on handcrafted heuristics.
\end{abstract}

\section{Introduction}

Variational autoencoders (VAEs) \citep{kingma2013auto} were originally designed for modeling continuous representations; however, applying them to discrete latent variable models is challenging due to non-differentiability. A common solution is to use gradient estimators tailored for discrete latent variables. The REINFORCE estimator \citep{williams1992simple} is an early example, providing an unbiased estimate of the gradient but suffering from high variance. Alternatively, methods such as the Gumbel-Softmax reparameterization trick \citep{jang2017categorical, maddison2017the} allow for a continuous relaxation of categorical distributions. While these methods introduce bias into the gradient estimation, they offer the benefit of significantly lower variance, thereby improving training stability.

Vector Quantized Variational Autoencoders (VQ-VAEs) \citep{van2017neural} extend the VAE framework to discrete latent spaces by discretizing continuous representations through a codebook via straight-through estimator (STE) \citep{bengio2013estimating}. Beyond the inherent variance-bias tradeoff in gradient estimation, VQ-VAEs are known to suffer from codebook collapse, wherein all encodings converge to a limited set of embedding vectors, resulting in the underutilization of many vectors in the codebook. This phenomenon diminishes the information capacity of the bottleneck. \citet{takida2022sq, williams2020hierarchical} hypothesized that deterministic quantization is the cause of codebook collapse and introduced stochastic sampling, leading to a entropy term in the log-likelihood lower bound. While high entropy is generally beneficial, it is inherently incompatible with Gumbel-Softmax gradient estimation. Various handcrafted heuristics have been proposed to mitigate this issue, including batch data-dependent k-means \citep{lancucki2020robust}, replacement policies \citep{zeghidour2021soundstream, dhariwal2020jukebox}, affine parameterization with alternate optimization \citep{huh2023straightening}, and entropy penalties \citep{yu2023language}. However, as these heuristics do not derive from the evidence lower Bound (ELBO), they cannot be unified within the variational Bayesian framework, rendering them ad-hoc solutions lacking a coherent foundation in variational inference.

\begin{figure}
    \centering
    \includegraphics[width=\textwidth]{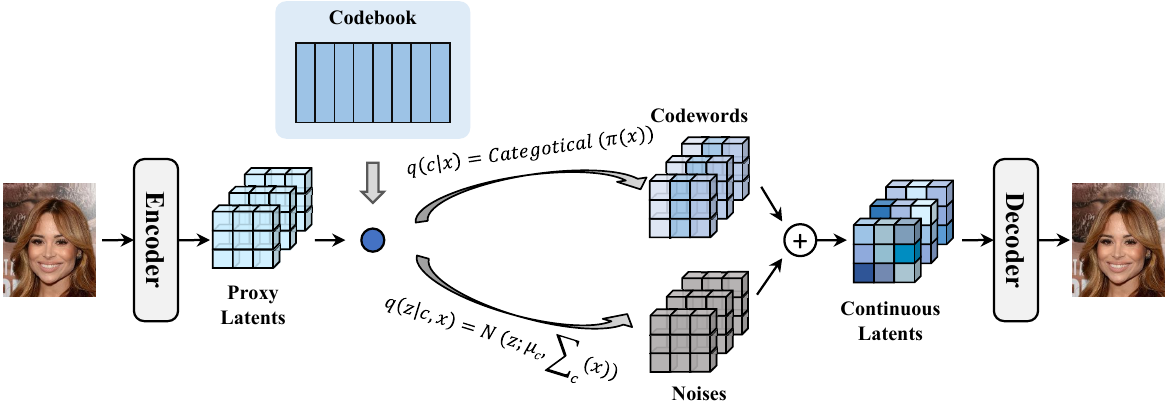}
    \caption{\textbf{Overview of GM-VQ}. First, the encoder deterministically maps the input to proxy latents, which are then used to retrieve corresponding codewords from the codebook and generate noise. The codewords and noise are then combined to form the continuous latents. Finally, these continuous latents are passed through the decoder to produce the final output.}
    \label{fig:diagram}
\end{figure}

In our work, we propose a Gaussian mixture prior based on VQ-VAE within a variational Bayesian framework, namely \textit{Gaussian Mixture Vector Quantization} (GM-VQ, see Figure \ref{fig:diagram}), combining the benefits of both discrete and continuous representations while avoiding handcrafted heuristics and strong assumptions. Additionally, to optimize the model and ensure compatibility with the gradient estimation errors inherent to Gumbel-Softmax, we modify the ELBO by replacing the conditional categorical posterior with an aggregated categorical posterior, resulting in an novel lower bound, the \textit{Aggregated Categorical Posterior Evidence Lower Bound} (ALBO), which minimizes estimation error while preserving codebook utilization. Concretely, our contributions are as follows:

\begin{itemize}
    \item To the best of our knowledge, we are the first to apply the Gaussian mixture prior formulation on VQ-VAE with strict adherence to the variational Bayesian framework.
    
    \item We introduce \textit{Aggregated Categorical Posterior Evidence Lower Bound} (ALBO), which is explicitly designed to be compatible with Gumbel-Softmax gradient estimation.
    \item We conduct experiments demonstrating improved codebook utilization and reduced information loss without relying on handcrafted heuristics.
\end{itemize}

\section{Preliminaries}

\subsection{Deterministic VQ-VAE}

The VQ-VAE \citep{van2017neural}, based on the VAE framework \citep{kingma2013auto}, learns discrete latent representations via vector quantization \citep{1162229}. For simplicity, we represent discrete latents with a single random variable \( \hat {\rvz} \) here, though in practice, we extract latent features of various dimensions.

Given an encoder output \( \hat{\rvz} \) from a high-dimensional input \( \rvx \) and a latent embedding space (codebook) \( \mathbf{M} \in \mathbb{R}^{C \times L} \), composed of \( C \) row vectors \( \boldsymbol{\mu}_{i} \in \mathbb{R}^L \), selects the discrete latent variable (codeword) as \( j = \arg \min_{i} \| \hat{\rvz} - \boldsymbol{\mu}_{i} \| \), yielding the approximate posterior distribution \( q(\rc \mid \rvx) \in \mathbb{R}^{C} \):

\begin{equation}
\label{eq:vqvae_posterior}
    q(\rc \mid \rvx) = [\1_{\mathrm{i = j}}]_{i=1}^{C} = \begin{cases}
        1, & \text{if } i = \arg \min_{i} \| \hat{\rvz} - \boldsymbol{\mu}_{i} \|, \\
        0, & \text{otherwise}.
    \end{cases}
\end{equation}

Based on the one-hot distribution \( q(\rc \mid \rvx) \), we have query vector \( \rvc_q = q(\rc \mid \rvx) \) for codebook, the quantized latent representation is expressed as \( \rvz_{c} = \rvc^T_q \mathbf{M} = \boldsymbol{\mu}_{j} \in \mathbb{R}^L \).

To handle the non-differentiability of the quantization process, the straight-through estimator (STE) \citep{bengio2013estimating} is applied, with the assumption \( \frac{\partial \rvz_{c}}{\partial \hat \rvz} = \mI \) to allow gradient flow. This assumption holds reasonably well when \( \rvz_{c} \) does not deviate significantly from \( \hat \rvz \); however, greater deviations introduce increased bias. To mitigate this issue and enhance gradient estimation, an additional discretization loss was introduced:

\begin{equation}
    \mathcal{L}_{\text{discretization}}(\hat \rvz, \rvz_{c}) = \| \hat \rvz - \text{sg}[\rvz_{c}] \|^2 + \alpha \cdot \| \text{sg}[\hat \rvz] - \rvz_{c} \|^2.
\end{equation}

Here, \( \text{sg}[\cdot] \) represents the stop-gradient operator, and \( \alpha \) adjusts the balance between minimizing the discrepancies between \( \hat \rvz \) and \( \rvz_q \).

\subsection{Stochastic VQ-VAE}
One problem of VQ-VAE is that the learned discrete representation uses only a fraction of
the full capacity of the codebook, a problem known
as codebook collapse. To solve this problem, stochastic sampling was introduced by modifying the approximate posterior , shifting from a one-hot representation to a distribution proportional to the negative squared distance between the encoder output \( \hat{\rvz} \) and the codewords \( \boldsymbol{\mu}_c \)(\citet{roy2018theory, sonderby2017continuous, shu2017compressing}):

\begin{equation}
q(\rc \mid \rvx) = \operatorname{Softmax}\left( -\frac{\| \hat{\rvz} - \boldsymbol{\mu}_c \|^2}{2\sigma^2} \right).
\end{equation}

To estimate gradients, the Gumbel-Softmax trick \citep{jang2017categorical, maddison2017the} approximates the categorical distribution as \( q^{g}(\rc \mid \rvx) = \operatorname{Softmax}_\tau (\log q_{i}(\rc \mid \rvx) + \rvg_i) \), where \( \rvg_i \) are independent and identically distributed (i.i.d.) samples from Gumbel(0,1). The index \( j \) is selected as \( j = \arg \max_i q_{i}^{g}(\rc \mid \rvx) \), resulting in the quantized representation \( \rvc_q = [\1_{i=j}]_{i=1}^C \).

In this framework, the gradient assumption shifts from \( \frac{\partial \rvz_c}{\partial \hat{\rvz}} = \displaystyle \mI \) to \( \frac{\partial \rvc_q}{\partial q^{g}(\rc \mid \rvx)} = \displaystyle \mI \). This gradient estimation requires a low entropy in \( q(\rc \mid \rvx) \) to be accurate . When entropy  \( q(\rc \mid \rvx) \) has a higher entropy the output distribution becomes more uniform. In this scenario, the Gumbel-Softmax trick outputs values that are less peaked, leading to noisy estimates for gradient-based optimization because the model becomes less certain about which category is the most likely. This uncertainty increases the inaccuracy in the gradient estimates. Thus, with a fixed temperature, high entropy in \( q(\rc \mid \rvx) \) leads to gradient estimation errors. However, this leads to a problem.

Assuming a uniform prior, the variational bound is:
\begin{equation}
    -\log p(\rvx) \leq \mathbb{E}_{\rc \sim q(\rc \mid \rvx)} \left[-\log p(\rvx \mid \rc)\right] - H(q(\rc \mid \rvx)) + \log C,
\end{equation}

which introduces a negative entropy loss that promotes a high-entropy posterior distribution, conflicting with low entropy requirement in Gumbel-Softmax trick and resulting in incompatible high gradient estimation bias.

\section{GM-VQ: Gaussian Mixture Vector Quantization}

To solve the problem mentioned above, we propose a stochastic vector quantization method using using Gaussian mixture distribution, which we refer to as \textit{Gaussian Mixture Vector Quantization} (GM-VQ), with the following encoding and decoding process:

\begin{equation}
    \rvx \xrightarrow{E_{\boldsymbol{\theta}}} \hat{\rvz} \xrightarrow{\mathbf{\mathbf{M}}} \rc \xrightarrow{\boldsymbol{\varepsilon}} \rvz \xrightarrow{D_{\boldsymbol{\phi}}} \tilde{\rvx}
\end{equation}

\begin{figure}
    \centering
    \includegraphics[width=0.4\textwidth]{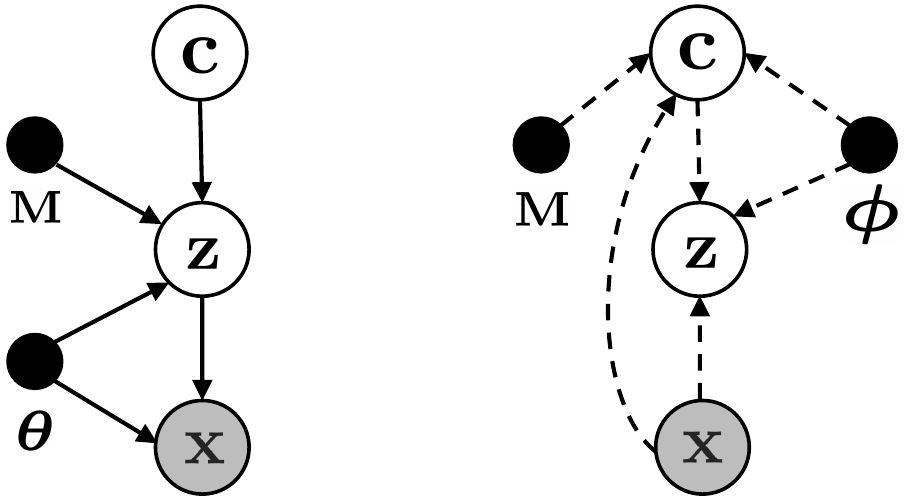}
    \caption{Probabilistic Graphical Model depicting the Gaussian Mixture Vector Quantization (GM-VQ) for the generative model (left) and the inference model (right).  The codebook \( \mathbf{M} \) plays a dual role, being shared between both the generative and inference models.}
    \label{fig:pgm}
\end{figure}

In our framework, GM-VQ comprises an encoder \( E_{\boldsymbol{\theta}} \) and a decoder \( D_{\boldsymbol{\phi}} \), parameterized by the deep neural network weights \( \boldsymbol{\theta} \) and \( \boldsymbol{\phi} \), respectively, and interconnected via a codebook \( \mathbf{M} \), which contains the means of Gaussian mixtures. The term \( \boldsymbol{\varepsilon} \) denotes the Gaussian noise introduced into the process. 

Unlike previous works adapted VQ-VAE with Gaussian mixture priors \citep{takida2022sq, williams2020hierarchical}, which directly transmit categorical representations to the decoder, our method explicitly adds noise to the categorical variables during training, feeding continuous latent variables into the decoder. While \citet{liu2021discrete} theoretically demonstrate that discrete representations can enhance generalization and robustness during evaluation, the codewords in our codebook are continuously updated throughout training, leading to a non-static categorical representation. By explicitly introducing noise, our method allows the decoder to adapt in parallel with the evolving codebook.

\subsection{Generative Model}

The generative model is defined by the joint distribution:

\begin{equation}
        p(\rvx, \rvz, \rc) = p(\rvx \mid \rvz) \, p(\rvz \mid \rc) \, p(\rc),
\end{equation}

where \( \rvx \in \mathbb{R}^D \) represents the observed data, \( \rvz \in \mathbb{R}^L \) is a continuous latent variable, and \( \rc \in \{1, 2, \ldots, C \} \) is a discrete latent variable indicating the mixture component. The latent variable \( \rvz \) follows a Gaussian mixture distribution, with the means stored in a codebook \( \mathbf{M} \in \mathbb{R}^{C \times L} \), where each row \( \boldsymbol{\mu}_c \) represents a codeword for component \( c \).

First, the discrete variable \( \rc \) is sampled from a categorical distribution, where \( \boldsymbol{\pi} \) represents the prior probabilities of the mixture components, typically assumed to be uniformly distributed. Then, given \( \rc \), the continuous latent variable \( \rvz \) is drawn from a multivariate Gaussian distribution with mean \( \boldsymbol{\mu}_c \) and isotropic covariance matrix \( \sigma_{\rvz}^2 \mI \). As \( \sigma_{\rvz}^2 \) approaches zero, this setup converges to deterministic quantization, similar to VQ-VAE.

Finally, the observed data \( \rvx \) is generated conditionally on \( \rvz \), where the decoder \( D_{\boldsymbol{\theta}} \) maps the continuous latents \( \rvz \) to the mean of the observation distribution, modeled as a Gaussian with fixed variance \( \sigma_{\rx}^2 \mI \).

\subsection{Variational Inference and Posterior Estimation}

To approximate the intractable posterior \( p(\rc, \rvz \mid \rvx) \), we employ a variational posterior \( q(\rc, \rvz \mid \rvx) \) following standard variational autoencoder methods. Without simplifying assumptions, we employ variational inference with an approximate posterior. We use the chain rule to factorize the posterior:

\begin{equation}
    q(\rc, \rvz \mid \rvx) = q(\rc \mid \rvx) q(\rvz \mid \rvx, \rc),
\end{equation}

allowing for conditional dependencies between \( \rvz \) and \( \rc \) given \( \rvx \). To achieve this, we introduce the following variational distributions:

\begin{enumerate}
    \item \textbf{Variational Distribution over \( \rc \)}:

    \begin{equation}
    q(\rc \mid \rvx) = \text{Categorical}\big( \boldsymbol{\pi}(\rvx) \big),
    \end{equation}

    where \( \boldsymbol{\pi}(\rvx) = (\boldsymbol{\pi}_1(\rvx), \boldsymbol{\pi}_2(\rvx), \ldots, \boldsymbol{\pi}_C(\rvx)) \) are the posterior probabilities over the mixture components, parameterized by an encoder \( E_{\boldsymbol{\theta}} \) and codebook \( \bm{M} \).

    \item \textbf{Variational Distribution over \( \rvz \)}:    

    We model \( q(\rvz \mid \rvx, \rc) \) as a multivariate Gaussian distribution centered at the codeword \( \boldsymbol{\mu}_c \), with a covariance \( \mathbf{\Sigma}_{c}(\rvx) \) that depends on \( \rvx \):

    \begin{equation}
    q(\rvz \mid \rvx, \rc) = \mathcal{N}\big( \rvz; \boldsymbol{\mu}_c, \mathbf{\Sigma}_{c}(\rvx) \big),  
    \end{equation}


    where \( \mathbf{\Sigma}_{c}(\rvx) = \sigma^2_{c}(\rvx) \mI \) is an isotropic covariance matrix. 
    
\end{enumerate}

\subsection{Training Objective and Optimization}

\begin{figure}
    \begin{minipage}{0.65\textwidth}
        \centering
        \includegraphics[width=\textwidth]{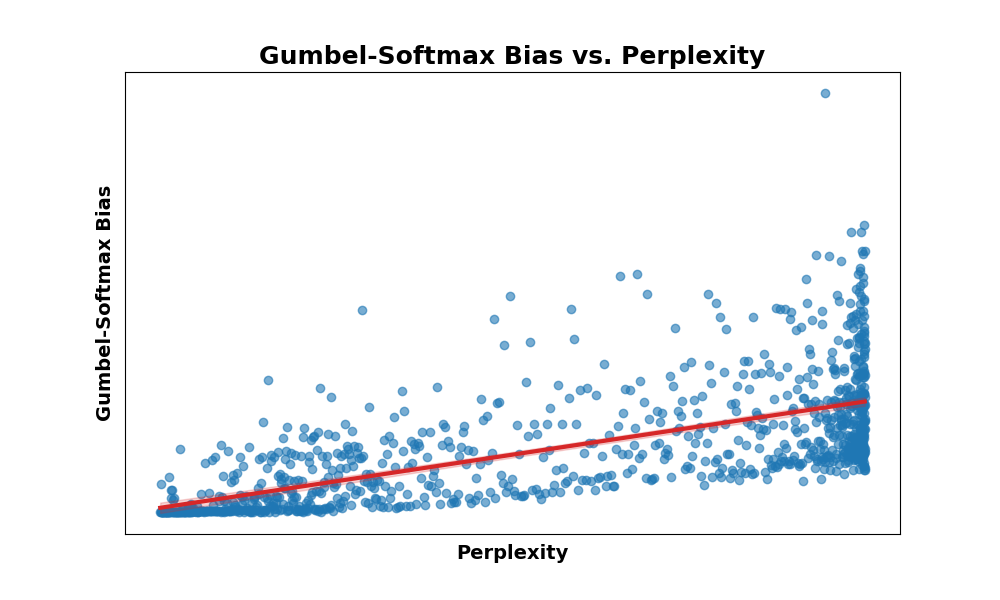}
    \end{minipage}\hfill
    \begin{minipage}{0.4\textwidth}
        Given a non-linear network, unnormalized logits corresponding to varying entropy levels are fed into the model, and the bias between Gumbel-Softmax gradient estimation and the exact gradient was calculated. A strong Pearson correlation (\( \rho = 0.77, p \leq 0.001^{***} \)) shows that gradient estimation errors increase with rising entropy. For more implementation details, Appendix \ref{sec:entropy_bias_implementation}.
    \end{minipage}
    \caption{Gradient Bias vs. Entropy Relationship}
\label{fig:be_relation}
\end{figure}

To train the model, we typically maximize the ELBO using Gumbel-Softmax gradient estimation. However, the presence of entropy in ELBO leads to poor Gumbel-Softmax gradient estimation(see Figure \ref{fig:be_relation}). Given that entropy can be particularly detrimental to Gumbel-Softmax gradient estimation, it is critical to mitigate its impact during training.

To resolve this, \citet{yu2023language} introduced two contrasting entropy measures derived from mutual information \citep{krause2010discriminative}:

\begin{equation}
    \mathcal{L}_\text{entropy} = H(q(\rc \mid \rvx)) - H(\mathbb{E}_{\rvx \sim p(\rvx)} q(\rc \mid \rvx)).
\end{equation}

In the context of vector quantization, the first term of this loss reduces uncertainty when mapping observed data \( \rvx \) to \( \rc \) and the second term encourages more discrete latent variables to be used, which decreases the codebook collapse problem.

However, this approach remains an additional heuristic that cannot be fully incorporated into the variational Bayesian framework. To address this limitation, we propose an alternative ELBO formulation.

\subsubsection{Aggregated Categorical Posterior Evidence Lower Bound}

We introduce the \textit{Aggregated Categorical Posterior Evidence Lower Bound} (ALBO) as an alternative to the traditional ELBO:

\begin{equation}
\mathcal{E}_{\text{ALBO}}(x)
= \mathbb{E}_{q(\rc) \, q(\rvz \mid \rvx)} \left[ \log \frac{p(\rvx, \rvz, \rc)}{q(\rc)} \right]
\leq \log p(\rvx).
\end{equation}

The ALBO provides a lower bound on the log-likelihood \( \log p(\rvx) \) (for derivation, see Appendix \ref{sec:albo}).

Here, \( q(\rc) \) represents the marginal distribution over \( \rc \), induced by the data distribution \( p(\rvx) \) and the approximate posterior \( q(\rc \mid \rvx) \), defined as \( q(\rc) = \int q(\rc \mid \rvx) p(\rvx) \, d\rvx \). However, since the true data distribution \( p(\rvx) \) is not accessible in practice, we have to rely on a finite dataset instead. To balance computational complexity and accuracy, we use a \textbf{mini-batch approximation}. For a mini-batch \( \mathcal{B} \) of size \( |\mathcal{B}| \), the marginal distribution \( q(\rc) \) is approximated as:

\begin{equation}
    q^{(\mathcal{B})}(\rc) = \frac{1}{|\mathcal{B}|} \sum_{\rvx \in \mathcal{B}} q(\rc \mid \rvx).
\end{equation}

This mini-batch approximation is computationally efficient and well-suited for stochastic optimization in deep learning.

For simplicity, \( \sigma_{\rvx}^2 \) and \( \sigma_{\rvz}^2 \) are typically fixed and not treated as learnable parameters. Thus, the objective function is constructed by minimizing the negative ALBO, resulting in the GM-VQ loss \( \mathcal{L}_{\text{GM-VQ}}(\rvx) \):

\begin{equation}
\begin{aligned}
\mathcal{L}_{\text{GM-VQ}}(\rvx) 
= \mathbb{E}_{q(\rvz \mid \rvx)} \| \rvx - D_{\boldsymbol{\theta}}(\rvz) \|^2
+ \gamma \cdot \mathcal{L}_{\text{reg}}(\rvx),
\end{aligned}
\end{equation}

where the regularization term \( \mathcal{L}_{\text{reg}}(\rvx) \) is defined as:

\begin{equation}
    \mathcal{L}_{\text{reg}}(\rvx) = \mathbb{E}_{q(\rc) q(\rvz \mid \rvx)} \| \rvz - \boldsymbol{\mu}_{c} \|^2 + \beta \cdot \KL(q(c) \Vert p(c)),
\end{equation}

with \( \beta \) and \( \gamma \) as non-negative hyperparameters controlling the balance between reconstruction and regularization. Detailed derivation can be found in Appendix \ref{sec:gmvq_loss}.

This loss function consists of three main components:
\begin{itemize}
    \item \( \mathbb{E}_{q(\rvz \mid \rvx)} \| \rvx - D_{\boldsymbol{\theta}}(\rvz) \|^2 \): Reconstruction loss, standard in autoencoder frameworks.
    \item \( \mathbb{E}_{q(\rc) q(\rvz \mid \rvx)} \| \rvz - \boldsymbol{\mu}_{c} \|^2 \): Latent regularization, ensures alignment between the latent variables and the learned codewords \( \boldsymbol{\mu}_c \).
    \item \( \KL(q(\rc) \Vert p(\rc)) \): Entropy term, enhances overall codebook utilization and prevents individual \( q(\rc \mid \rvx) \) from drifting towards high entropy, thereby reducing gradient estimation errors.
\end{itemize}

\subsubsection{Parameterization of the Variational Distributions}

\paragraph{Variational Categorical Distribution \( q(\rc \mid \rvx) \)}

To compute \( q(\rc \mid \rvx) \), we use the encoder \( E_{\boldsymbol{\phi}} \) to generate a proxy representation \( \hat{\rvz}(\rvx) \in \mathbb{R}^{L} \) and raw weights \( \hat{\rvr}(\rvx) \in \mathbb{R}^{L} \), which are then activated into positive weights \( \hat{\rvw}(\rvx) \) using the Softplus function \( \zeta(\cdot) \):

\begin{equation}
\hat{\rvz}(\rvx), \, \hat{\rvr}(\rvx) = E_{\boldsymbol{\phi}}(\rvx), \quad \hat{\rvw}(\rvx) = \zeta\big( \hat{\rvr}(\rvx) \big).
\end{equation}

The unnormalized log probabilities \( \vl_c(\rvx) \) are then computed based on the Mahalanobis-like distance between \( \hat{\rvz}(\rvx) \) and each codeword \( \boldsymbol{\mu}_c \):

\begin{equation}
\begin{aligned}
\vl_c(\rvx)
& = -\frac{1}{2} \left( \hat{\rvz}(\rvx) - \boldsymbol{\mu}_c \right)^\top \text{diag}(\hat{\mathbf{w}}(\rvx)) \left( \hat{\rvz}(\rvx) - \boldsymbol{\mu}_c \right)
\\
& = -\frac{1}{2} \sum_{i=1}^{L} \hat{\rvw}_i(\rvx) \left( \hat{\rvz}_i(\rvx) - \boldsymbol{\mu}_{c,i} \right)^2.  
\end{aligned}
\end{equation}

The posterior probabilities \( \boldsymbol{\pi}_c(\rvx) \) are obtained via softmax:

\begin{equation}
\boldsymbol{\pi}_c(\rvx)
= \operatorname{Softmax}\big( \vl_c(\rvx) \big)
= \frac{\exp\big( \vl_c(\rvx) \big)}{\sum_{c'=1}^C \exp\big( \vl_{c'}(x) \big)}.   
\end{equation}

This formulation ensures that components with codewords closer to \( \hat{\rvz}(\rvx) \) have higher posterior probabilities.

\paragraph{Variational Continuous Distribution \( q(\rvz \mid \rvx, \rc) \)}

The variance \( \boldsymbol{\sigma}^2_{\rc}(\rvx) \) in \( q(\rvz \mid \rvx, \rc) \) reflects the uncertainty in assigning \( \rvx \) to component \( \rc \). It is parameterized based on the squared distance between the encoder's output \( \hat{\rvz}(\rvx) \) and the codeword \( \boldsymbol{\mu}_c \):

\begin{equation}
\boldsymbol{\sigma}^2_{c}(\rvx)
= \frac{\| \hat{\rvz}(\rvx) - \boldsymbol{\mu}_\rc \|^{2} / L}{2 \sigma^2},
\end{equation}

where \( \sigma^2 \) is the scalar variance from the generative model and \( L \) is the latent dimensionality.

This parameterization allows \( q(\rvz \mid \rvx, \rc) \) to adapt its variance based on the distance between \( \hat{\rvz}(\rvx) \) and \( \boldsymbol{\mu}_c \), ensuring a flexible representation of uncertainty. As proxy latent \( \hat{\rvz}(\rvx) \) approaches \( \boldsymbol{\mu}_c \), the variance \( \boldsymbol{\sigma}^2_{c}(\rvx) \) decreases, indicating higher confidence in the assignment to component \( c \). Conversely, when the distance increases, the variance grows, signaling greater uncertainty.

\paragraph{Reparameterization and Codebook Update}

To enable backpropagation through the discrete sampling of \( \rc \), we use the Gumbel-Softmax reparameterization trick. The discrete latent variable \( \rc \) is computed as \( j = \arg \max \, \operatorname{Softmax}_\tau(\log q(\rc \mid \rvx) + \rvg) \), where \( \rvg \) are i.i.d. Gumbel(0,1) samples, yielding the quantized representation \( \rvc_q = [\1_{i=j}]_{i=1}^C \).

For the continuous latent variable \( \rvz \), we apply the standard VAE reparameterization: \( \rvz = \boldsymbol{\mu}_c + \boldsymbol{\sigma}_c(\rvx) \odot \boldsymbol{\epsilon} \), with \( \boldsymbol{\epsilon} \sim \mathcal{N}(0, \mI) \). In the ALBO framework, we sample from \( q(\rvz \mid \rvx) \), combining the quantized codeword and noise as \( \rvz = \rvc_q^T \mathbf{M} + \boldsymbol{\sigma}_c(\rvx) \odot \boldsymbol{\epsilon} \).

This formulation ensures that all codewords \( \boldsymbol{\mu}_c \) are naturally updated during optimization, preventing the codebook collapse problem without the need for additional commitment loss functions or exponential moving averages. It enables the model to manage deviations from the codewords in a controlled and efficient manner.





\section{Related Work}

Variational Autoencoders (VAEs) \citep{kingma2013auto} comprise a generative model and a recognition model, bridged by a latent variable typically modeled with a multivariate Gaussian prior. While the generative component is well-known, the recognition model effectively learns continuous representations from data \citep{zhang2022improving, yang2021causalvae, zhao2017infovae, higgins2017beta}.


To address discrete representation learning, Vector Quantized VAE (VQ-VAE) \citep{razavi2019generating, van2017neural} employs vector quantization \citep{1162229} to discretize latent embeddings under a uniform prior. Since the quantization process is non-differentiable, techniques like the straight-through estimator \citep{bengio2013estimating} are used to approximate gradients, introducing potential bias. VQ-VAE also incorporates a discretization loss to mitigate these issues, but challenges such as codebook underutilization and information loss remain. \citet{sonderby2017continuous, shu2017compressing} introduced stochastic sampling based on the negative distance and applied Gumbel-Softmax for gradient estimation. Later, \citet{karpathy2021deepvectorquantization, esser2021taming} proposed an encoder that directly outputs the posterior, applying Gumbel-Softmax without conditioning on the codebook.

While many existing VAEs utilize Gaussian mixture priors \citep{liu2023cloud, bai2022gaussian, falck2021multi, guo2020variational, jiang2016variational, dilokthanakul2016deep, nalisnick2016approximate}, our approach is distinct in its close connection with vector quantization. Specifically, we reuse the means from the codebook, whereas in other GMM-based models, the posterior means are typically learned transiently, conditioned on different components or networks. Although \citet{takida2022sq, williams2020hierarchical} also explore Gaussian mixture priors in VQ-VAE, they modify the reconstruction loss and feed discrete latents directly into the decoder, deviating from strict adherence to the ELBO. Furthermore, prior works often rely on simplifying assumptions on variational posterior, such as mean-field \citep{liu2023cloud, falck2021multi, figueroa2017simple, jiang2016variational} or Markovian assumptions \citep{takida2022sq, williams2020hierarchical}.

Previous works \citep{tomczak2018vae, hoffman2016elbo, makhzani2015adversarial} have attempted to modify the ELBO framework by averaging the objective over data distribution, but they retain the original variational conditional distribution, leaving the entropy term of the conditional posterior it intact. In contrast, our approach is specifically motivated by the gradient estimation error in categorical latents by using an aggregated categorical posterior instead of the conditional categorical posterior.

Beyond these theoretical advancements, vector quantization (VQ) and its related concepts have also seen extensive applications across various domains. Here, we highlight some recent works. UniMoT \citep{zhang2024unimot} introduces a VQ-driven tokenizer that converts molecules into molecular token sequences, while VQSynergy \citep{wu2024vqsynery} integrates VQ for drug synergy prediction. In image and video generation, MAGVIT-v2 \citep{yu2023language} applies VQ-VAE to achieve high-fidelity reconstructions, and VAR \citep{tian2024visual} leverages VQ to advance visual autoregressive learning. These studies demonstrate the pivotal role of VQ in both theoretical advancements and practical applications.

\section{Experiments}

In this section, we provide a comprehensive analysis of our experiments using the proposed GM-VQ model, focusing on its performance in image reconstruction tasks across two benchmark datasets CIFAR10 and CelebA.

\subsection{Experimental Setup}

We use Gumbel-Softmax for gradient estimation throughout our experiments, adopting an annealing schedule similar to \citet{takida2022sq, jang2017categorical}. The temperature starts at 2.0 and is gradually reduced to 0.1 during training.

Our architecture and hyperparameters closely follow the setup in \citet{huh2023straightening}. For the CIFAR10 dataset (32x32 image size), we use a compact architecture consisting of convolutional layers followed by vector quantization, similar to standard autoencoders. For CelebA (resized to 128x128), a deeper network is employed to manage the higher resolution. Both architectures incorporate a VQ layer to quantize the latent representations into discrete codes.

Models are trained for 100 epochs on both datasets, using a batch size of 256. We employ AdamW \citep{loshchilov2017decoupled} as the optimizer, with a maximum learning rate of 1e-2 for CIFAR10 and 1e-4 for CelebA. The learning rate follows a cosine decay schedule with linear warmup as in \citet{huh2023straightening}, with 10 epochs of warmup starting at a factor of 0.2, followed by 90 epochs of cosine decay. All models are initialized using K-means clustering for codebook initialization, following \citet{esser2021taming}. We use 1024 codes with a latent dimension of 64 across all experiments, applying the same weight decay to both the encoder-decoder and the codebook. During evaluation, no noise is added to the reconstructed \( \rvx \) for maximum likelihood estimation, and latents \( \rvz \) are directly sampled from the codebook \( \mathbf{M} \) without extra noise.

We report Mean Squared Error (MSE) as the metric for reconstruction quality, which measures the mean pixel-wise difference between the original and reconstructed images. Additionally, we report perplexity, defined as \( 2^{H(q)} \), to evaluate the diversity of codebook usage. Higher perplexity indicates a more balanced use of the available codes. Notably, this perplexity is not based on the entropy of individual codes \( q(\rc \mid \rvx) \) but on the average entropy across a batch of categorical distributions \( q^{(\mathcal{B})}(\rc) \), with perplexity computed per batch and then averaged across all batches.

We compare our GM-VQ model against several baseline methods commonly used in vector quantization-based image reconstruction. The primary baseline is the standard VQ-VAE \citep{van2017neural}. Variants include VQ-VAE + $l_2$ \citep{yu2021vector}, which stabilizes training through $l_2$ normalization, and VQ-VAE + replace \citep{zeghidour2021soundstream, dhariwal2020jukebox}, which replaces unused code vectors with random embeddings to avoid codebook collapse. SQ-VAE \citep{takida2022sq} introduces stochastic quantization for improved code diversity, while Gumbel-VQVAE \citep{karpathy2021deepvectorquantization, esser2021taming} employs Gumbel-softmax for smoother gradient updates. Lastly, VQ-VAE + affine + OPT \citep{huh2023straightening} addresses codebook covariate shift with affine parameterization and alternating training. Our primary model, GM-VQ, was tuned by fixing \( \beta = 1 \) and selecting the best \( \gamma \), while the variant GM-VQ + Entropy was tuned with higher entropy regularization (\( \beta > 1 \)) and the fixed \( \gamma \), promoting more balanced codebook usage.

\subsection{Performance Comparison}

\begin{table}[htbp!]
    \centering
    \scalebox{0.9}{
    \begin{tabular}{lcccc}
        \toprule
        \textbf{Method} & \multicolumn{2}{c}{\textbf{CIFAR10}} & \multicolumn{2}{c}{\textbf{CELEBA}} \\
        \cmidrule(lr){2-3} \cmidrule(lr){4-5}
        & \textbf{MSE (10$^{-3}$) $\downarrow$} & \textbf{Perplexity $\uparrow$} & \textbf{MSE (10$^{-3}$) $\downarrow$} & \textbf{Perplexity $\uparrow$} \\
        \midrule
        VQVAE & 5.65 & 14.0 & 10.02 & 16.2 \\
        VQVAE + $l_2$ & 3.21 & 57.0 & 6.49 & 188.7 \\
        VQVAE + replace & 4.07 & 109.8 & 4.77 & 676.4 \\
        VQVAE + $l_2$ + replace & 3.24 & 115.6 & 4.93 & 861.7 \\
        VQVAE + Affine & 5.15 & 69.5 & 7.47 & 112.6 \\
        VQVAE + OPT & 4.73 & 15.5 & 7.78 & 30.5 \\
        VQVAE + Affine + OPT & 4.00 & 79.3 & 6.60 & 186.6 \\
        SQVAE & 3.36 & 769.3 & 9.17 & 769.1 \\
        Gumbel-VQVAE & 6.16 & 20.3 & 7.34 & 96.7 \\
        \midrule
        GM-VQ & 3.13 & 731.9 & 1.38 & 338.6 \\
        GM-VQ + Entropy & 3.11 & 878.7 & 0.97 & 831.0 \\
        \bottomrule
    \end{tabular}}
    \caption{Comparison of methods on CIFAR10 and CELEBA datasets using MSE and Perplexity metrics.}
    \label{table:res}
\end{table}

We evaluate the performance of GM-VQ on the CIFAR10 and CelebA datasets, comparing it against several baseline methods. Table \ref{table:res} presents the results, using MSE for reconstruction accuracy and perplexity metrics.

In the CIFAR10 dataset, GM-VQ achieves an MSE of 3.13, a significant improvement over the standard VQVAE (MSE 5.65) and variants like VQVAE + $l_2$ (MSE 3.21) and VQVAE + replace (MSE 4.07). In terms of codebook utilization, GM-VQ achieves a perplexity of 731.9, considerably higher than VQ-VAE + replace (perplexity 109.8), indicating more efficient and diverse code usage. This highlights GM-VQ's ability to mitigate codebook collapse and ensure robust code assignments.

On the CelebA dataset, GM-VQ excels with an MSE of 1.38, significantly outperforming the baseline VQVAE (MSE 10.02) and VQVAE + replace (MSE 4.77). GM-VQ also maintains strong codebook diversity with a perplexity score of 338.6. The GM-VQ + Entropy variant further enhances performance, achieving the lowest MSE of 0.97 and a perplexity of 831.0. This shows that entropy regularization effectively promotes balanced codebook usage without sacrificing reconstruction quality.

In summary, across both datasets, GM-VQ and GM-VQ + Entropy consistently outperform all baseline models in terms of both reconstruction accuracy and codebook utilization. These results demonstrate the model's robustness and its ability to maintain a balance between reconstruction fidelity and efficient code usage.

\subsection{Impact of Entropy Regularization}

\begin{figure}
    \centering
    \includegraphics[width=0.7\textwidth]{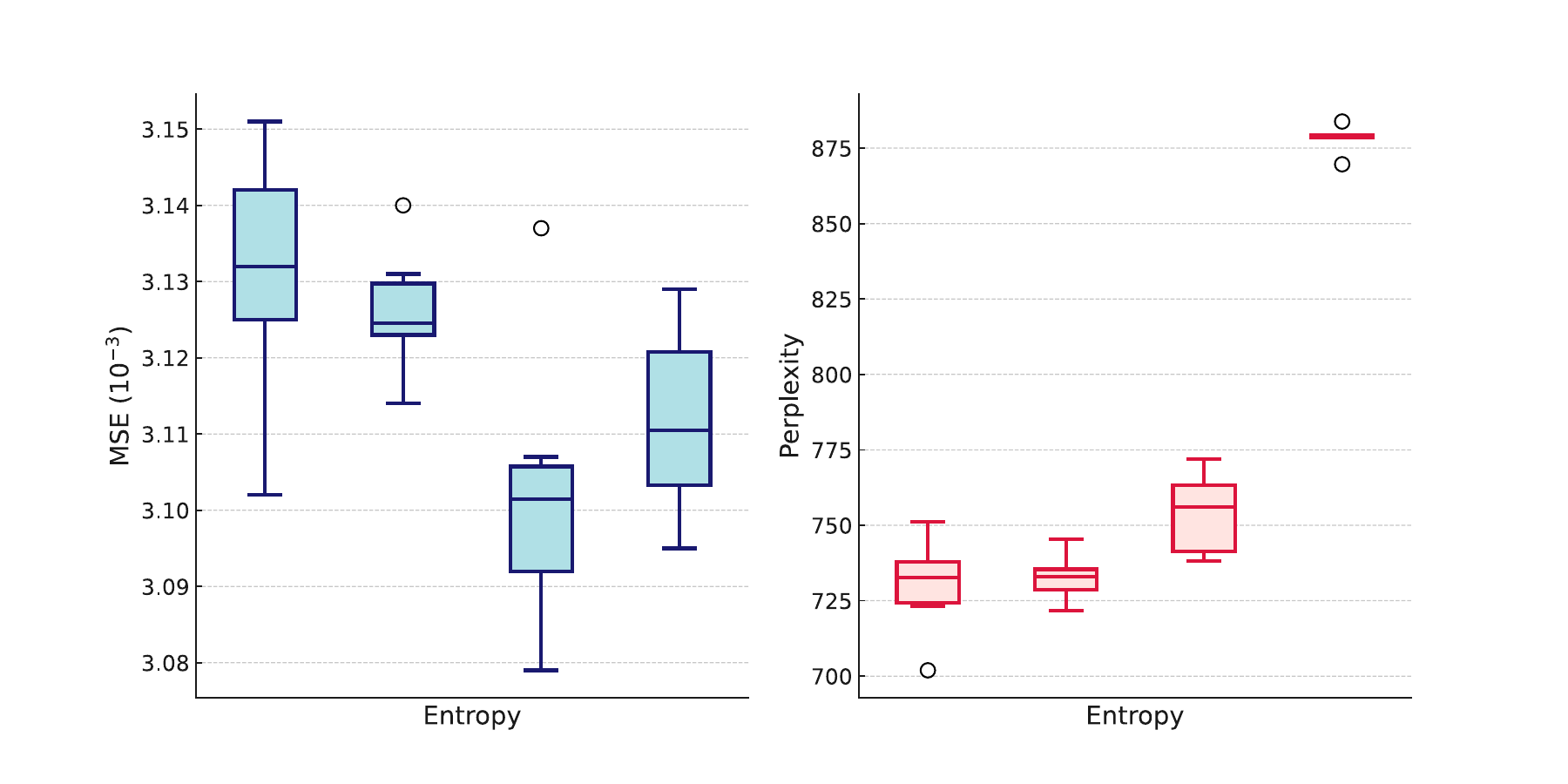}
    \caption{Box plots showing the impact of entropy regularization on reconstruction quality (MSE) and codebook utilization (Perplexity) for the GM-VQ model. The left panel demonstrates the general trend of decreasing MSE with increasing entropy, the right pane shows the rise in perplexity with higher perplexity.}
    \label{fig:boxplot_mse_ppl}
\end{figure}


To further demonstrate the compatibility of the aggregated posterior with increased entropy, we conducted experiments across different entropy regularization levels using GM-VQ on the CIFAR10 dataset.

The box plots in Figure \ref{fig:boxplot_mse_ppl} provide a summary of how changes in entropy affect both Mean Squared Error (MSE) and Perplexity. In the left panel, we observe a general trend where MSE decreases as entropy increases, indicating a tendency toward improved reconstruction quality, though not uniformly across all entropy levels. Meanwhile, the right panel shows that higher entropy promotes more effective codebook usage, with perplexity rising as entropy grows, reflecting more balanced code assignments.

This trend suggests that while increased entropy yields better codebook utilization (higher perplexity), it also drives improvements in reconstruction accuracy (lower MSE).

\section{Conclusion}

In summary, the GM-VQ framework extends the traditional VQ-VAE by incorporating a probabilistic structure grounded in a Gaussian mixture model. By employing the ALBO objective, we ensure optimization is well-suited to Gumbel-Softmax gradient estimation. Our empirical results demonstrate the effectiveness of GM-VQ, highlighting its ability to achieve a strong balance between reconstruction quality and codebook diversity.

\section*{Acknowledgments}

We sincerely thank Michael Mozer for his valuable suggestions on the idea, as well as his careful revisions and insightful improvements to the manuscript. We are also grateful to Wenhao Zhao for providing experimental results that served as helpful references.

\newpage

\newpage
\bibliography{iclr2025_conference}
\bibliographystyle{iclr2025_conference}

\appendix
\section{Appendix}

\subsection{Implementation Details of Entropy Bias}
\label{sec:entropy_bias_implementation}

To investigate the relationship between entropy and the bias in Gumbel-Softmax gradient estimation, we conducted targeted experiments.

We used a multi-layer perceptron (MLP) with hidden layers of size [50, 5], where the input consists of 10 possible actions and the output is a scalar. To obtain the exact gradients, we pass each categorical one-hot action through the non-linear decoder. For the Gumbel-Softmax gradient estimation, we repeated the experiment 50 times to compute the empirical average of the estimated gradients.

To assess the impact of entropy on bias, we varied the input entropy by applying softmax with different temperature (\(\tau\)) values to a fixed set of unnormalized logits, which determined the input probabilities for the MLP. This setup allowed us to analyze how changes in entropy influence the bias in Gumbel-Softmax gradient estimation.

\subsection{Derivation of ALBO}
\label{sec:albo}

The \textit{Aggregated Categorical Posterior Evidence Lower Bound} (ALBO) provides a lower bound on the log-likelihood \( \log p(\rvx) \), similar in structure to the Evidence Lower Bound (ELBO) commonly used in variational inference. This bound is derived through the application of Jensen's inequality, as shown below:

\begin{equation}
\begin{aligned}
\log p(\rvx)
&= \log \mathbb{E}_{q(\rc) q(\rvz \mid \rvx)} \left[ \frac{p(\rvx, \rvz, \rc)}{q(\rc) q(\rvz \mid \rvx)} \right] \\
&\geq \mathbb{E}_{q(\rc) q(\rvz \mid \rvx)} \left[ \log \frac{p(\rvx, \rvz, \rc)}{q(\rc) q(\rvz \mid \rvx)} \right] \quad \text{(by Jensen's inequality)} \\
&= \mathbb{E}_{q(\rc) q(\rvz \mid \rvx)} \left[ \log \frac{p(\rvx, \rvz, \rc)}{q(\rc)} \right] - \mathbb{E}_{q(\rvz \mid \rvx)} \left[ \log q(\rvz \mid \rvx) \right] \\
&\geq \mathbb{E}_{q(\rc) q(\rvz \mid \rvx)} \left[ \log \frac{p(\rvx, \rvz, \rc)}{q(\rc)} \right] \quad \text{(since } \mathbb{E}_{q(\rvz \mid \rvx)} \left[ \log q(\rvz \mid \rvx) \right] \leq 0 \text{)} \\
&= \mathcal{E}_{\text{ALBO}}(\rvx).
\end{aligned}  
\end{equation}

By applying Jensen's inequality, the logarithm is moved inside the expectation, yielding a tractable lower bound. The term \( \mathbb{E}_{q(\rvz \mid \rvx)} \left[ \log q(\rvz \mid \rvx) \right] \) represents the entropy of the posterior over \( \rvz \). Since entropy is non-positive, it further tightens the bound, ensuring that \( \mathcal{E}_{\text{ALBO}}(\rvx) \) provides a meaningful approximation.

Thus, \( \mathcal{E}_{\text{ALBO}}(\rvx) \) serves as a valid lower bound on \( \log p(\rvx) \), similar to the ELBO in traditional variational inference. The key difference is in the aggregation of the categorical posterior over \( \rc \), which offers better compatibility with probability-based gradient estimation.

\subsection{Derivation of GM-VQ Loss}
\label{sec:gmvq_loss}

Minimizing \( \mathcal{L}_{\text{GM-VQ}}(\rvx) \) ensures that the model effectively reconstructs the data while regularizing the latent distributions to align with the priors, thereby improving generalization. Below, we provide the detailed derivation of the GM-VQ loss.

We start by maximizing the \( \mathcal{E}_{\text{ALBO}}(\rvx) \):

\begin{equation}
\begin{aligned}
\underset{\boldsymbol{\theta}, \boldsymbol{\phi}, \mathbf{M}}{\arg \max} \, &
\mathcal{E}_{\text{ALBO}}(\rvx)
\\
= \underset{\boldsymbol{\theta}, \boldsymbol{\phi}, \mathbf{M}}{\arg \min} \, &
- \mathcal{E}_{\text{ALBO}}(\rvx)
\\
= \underset{\boldsymbol{\theta}, \boldsymbol{\phi}, \mathbf{M}}{\arg \min} \, &
- \mathbb{E}_{q(\rc) q(\rvz \mid \rvx)} \left[ \log \frac{p(\rvx, \rvz, \rc)}{q(\rc)} \right] \\
= \underset{\boldsymbol{\theta}, \boldsymbol{\phi}, \mathbf{M}}{\arg \min} \, &
\mathbb{E}_{q(\rc) q(\rvz \mid \rvx)} \left[ -\log \frac{p(\rvx \mid \rvz) \, p(\rvz \mid \rc) \, p(\rc)}{q(\rc)} \right] \\
= \underset{\boldsymbol{\theta}, \boldsymbol{\phi}, \mathbf{M}}{\arg \min} \, &
\mathbb{E}_{q(\rvz \mid \rvx)} \left[ -\log p(\rvx \mid \rvz) \right]
+ \mathbb{E}_{q(\rc) q(\rvz \mid \rvx)} \left[ -\log p(\rvz \mid \rc) \right]
+ \KL(q(\rc) \Vert p(\rc))
\\
= \underset{\boldsymbol{\theta}, \boldsymbol{\phi}, \mathbf{M}}{\arg \min} \, &
\mathbb{E}_{q(\rvz \mid \rvx)} \left[ \frac{\| \rvx - D_{\boldsymbol{\theta}}(\rvz) \|^2}{2\sigma^2_{\rvx}} \right]
+ \mathbb{E}_{q(\rc) q(\rvz \mid \rvx)} \left[ \frac{\| \rvz - 
\boldsymbol{\mu}_{c} \|^2}{2\sigma_{\rvz}^2} \right]
+ \KL(q(c) \Vert p(c))
\\
= \underset{\boldsymbol{\theta}, \boldsymbol{\phi}, \mathbf{M}}{\arg \min} \, &
\mathbb{E}_{q(\rvz \mid \rvx)} \| \rvx - D_{\boldsymbol{\theta}}(\rvz) \|^2
+ \frac{\sigma^2_{\rvx}}{\sigma_{\rvz}^2} \cdot \left( \mathbb{E}_{q(\rc) q(\rvz \mid \rvx)}\| \rvz - \boldsymbol{\mu}_{c} \|^2
+ 2\sigma_{\rvz}^2 \cdot \KL(q(c) \Vert p(c)) \right)
\\
\end{aligned}
\end{equation}

Given variance terms \( \sigma_{\rvx}^2 \) and \( \sigma_{\rvz}^2 \) are fixed, we can replace them with positive hyperparameters \( \beta \) and \( \gamma \), respectively, to simplify this expression, where:
\begin{enumerate}
    \item \( \gamma = \frac{\sigma_{\rvx}^2}{\sigma_{\rvz}^2} \) controls the balance between reconstruction and latent regularization.
    \item \( \beta = 2\sigma_{\rvz}^2 \) modulates the strength of the KL divergence regularization.
\end{enumerate}

Thus, the GM-VQ loss can be rewritten as:

\begin{equation}
\begin{aligned}
\mathcal{L}_{\text{GM-VQ}}(\rvx) 
= \mathbb{E}_{q(\rvz \mid \rvx)} \| \rvx - D_{\boldsymbol{\theta}}(\rvz) \|^2
+ \gamma \cdot \left( \mathbb{E}_{q(\rc) q(\rvz \mid \rvx)} \| \rvz - \boldsymbol{\mu}_{c} \|^2 
+ \beta \cdot \KL(q(c) \Vert p(c)) \right)
\end{aligned}
\end{equation}

This formulation, with hyperparameters \( \beta \) and \( \gamma \), balances the reconstruction fidelity and the regularization of the latent space.

\end{document}

%% file: math_commands.tex
\usepackage{amsmath,amsfonts,bm}

\def\eqref#1{equation~\ref{#1}}

\def\1{\bm{1}}

\def\rc{{\textnormal{c}}}

\def\rx{{\textnormal{x}}}

\def\rvc{{\mathbf{c}}}

\def\rvg{{\mathbf{g}}}

\def\rvr{{\mathbf{r}}}

\def\rvw{{\mathbf{w}}}
\def\rvx{{\mathbf{x}}}

\def\rvz{{\mathbf{z}}}

\def\vl{{\bm{l}}}

\def\mI{{\bm{I}}}

\DeclareMathAlphabet{\mathsfit}{\encodingdefault}{\sfdefault}{m}{sl}
\SetMathAlphabet{\mathsfit}{bold}{\encodingdefault}{\sfdefault}{bx}{n}

\newcommand{\KL}{D_{\mathrm{KL}}}



%% file: main.bbl
\begin{thebibliography}{42}
\providecommand{\natexlab}[1]{#1}
\providecommand{\url}[1]{\texttt{#1}}
\expandafter\ifx\csname urlstyle\endcsname\relax
  \providecommand{\doi}[1]{doi: #1}\else
  \providecommand{\doi}{doi: \begingroup \urlstyle{rm}\Url}\fi

\bibitem[Bai et~al.(2022)Bai, Kong, and Gomes]{bai2022gaussian}
Junwen Bai, Shufeng Kong, and Carla~P Gomes.
\newblock Gaussian mixture variational autoencoder with contrastive learning for multi-label classification.
\newblock In \emph{international conference on machine learning}, pp.\  1383--1398. PMLR, 2022.

\bibitem[Bengio et~al.(2013)Bengio, L{\'e}onard, and Courville]{bengio2013estimating}
Yoshua Bengio, Nicholas L{\'e}onard, and Aaron Courville.
\newblock Estimating or propagating gradients through stochastic neurons for conditional computation.
\newblock \emph{arXiv preprint arXiv:1308.3432}, 2013.

\bibitem[Dhariwal et~al.(2020)Dhariwal, Jun, Payne, Kim, Radford, and Sutskever]{dhariwal2020jukebox}
Prafulla Dhariwal, Heewoo Jun, Christine Payne, Jong~Wook Kim, Alec Radford, and Ilya Sutskever.
\newblock Jukebox: A generative model for music.
\newblock \emph{arXiv preprint arXiv:2005.00341}, 2020.

\bibitem[Dilokthanakul et~al.(2016)Dilokthanakul, Mediano, Garnelo, Lee, Salimbeni, Arulkumaran, and Shanahan]{dilokthanakul2016deep}
Nat Dilokthanakul, Pedro~AM Mediano, Marta Garnelo, Matthew~CH Lee, Hugh Salimbeni, Kai Arulkumaran, and Murray Shanahan.
\newblock Deep unsupervised clustering with gaussian mixture variational autoencoders.
\newblock \emph{arXiv preprint arXiv:1611.02648}, 2016.

\bibitem[Esser et~al.(2021)Esser, Rombach, and Ommer]{esser2021taming}
Patrick Esser, Robin Rombach, and Bjorn Ommer.
\newblock Taming transformers for high-resolution image synthesis.
\newblock In \emph{Proceedings of the IEEE/CVF conference on computer vision and pattern recognition}, pp.\  12873--12883, 2021.

\bibitem[Falck et~al.(2021)Falck, Zhang, Willetts, Nicholson, Yau, and Holmes]{falck2021multi}
Fabian Falck, Haoting Zhang, Matthew Willetts, George Nicholson, Christopher Yau, and Chris~C Holmes.
\newblock Multi-facet clustering variational autoencoders.
\newblock \emph{Advances in Neural Information Processing Systems}, 34:\penalty0 8676--8690, 2021.

\bibitem[Figueroa(2017)]{figueroa2017simple}
Jhosimar~Arias Figueroa.
\newblock Is simple better?: Revisiting simple generative models for unsupervised clustering.
\newblock In \emph{NIPS Workshop on Bayesian Deep Learning}, 2017.

\bibitem[Gray(1984)]{1162229}
R.~Gray.
\newblock Vector quantization.
\newblock \emph{IEEE ASSP Magazine}, 1\penalty0 (2):\penalty0 4--29, 1984.
\newblock \doi{10.1109/MASSP.1984.1162229}.

\bibitem[Guo et~al.(2020)Guo, Zhou, Chen, Ying, Zhang, and Zhou]{guo2020variational}
Chunsheng Guo, Jialuo Zhou, Huahua Chen, Na~Ying, Jianwu Zhang, and Di~Zhou.
\newblock Variational autoencoder with optimizing gaussian mixture model priors.
\newblock \emph{IEEE Access}, 8:\penalty0 43992--44005, 2020.

\bibitem[Higgins et~al.(2017)Higgins, Matthey, Pal, Burgess, Glorot, Botvinick, Mohamed, and Lerchner]{higgins2017beta}
Irina Higgins, Loic Matthey, Arka Pal, Christopher~P Burgess, Xavier Glorot, Matthew~M Botvinick, Shakir Mohamed, and Alexander Lerchner.
\newblock beta-vae: Learning basic visual concepts with a constrained variational framework.
\newblock \emph{ICLR (Poster)}, 3, 2017.

\bibitem[Hoffman \& Johnson(2016)Hoffman and Johnson]{hoffman2016elbo}
Matthew~D Hoffman and Matthew~J Johnson.
\newblock Elbo surgery: yet another way to carve up the variational evidence lower bound.
\newblock In \emph{Workshop in Advances in Approximate Bayesian Inference, NIPS}, volume~1, 2016.

\bibitem[Huh et~al.(2023)Huh, Cheung, Agrawal, and Isola]{huh2023straightening}
Minyoung Huh, Brian Cheung, Pulkit Agrawal, and Phillip Isola.
\newblock Straightening out the straight-through estimator: Overcoming optimization challenges in vector quantized networks.
\newblock In \emph{International Conference on Machine Learning}, pp.\  14096--14113. PMLR, 2023.

\bibitem[Jang et~al.(2017)Jang, Gu, and Poole]{jang2017categorical}
Eric Jang, Shixiang Gu, and Ben Poole.
\newblock Categorical reparameterization with gumbel-softmax.
\newblock In \emph{International Conference on Learning Representations}, 2017.
\newblock URL \url{https://openreview.net/forum?id=rkE3y85ee}.

\bibitem[Jiang et~al.(2016)Jiang, Zheng, Tan, Tang, and Zhou]{jiang2016variational}
Zhuxi Jiang, Yin Zheng, Huachun Tan, Bangsheng Tang, and Hanning Zhou.
\newblock Variational deep embedding: An unsupervised and generative approach to clustering.
\newblock \emph{arXiv preprint arXiv:1611.05148}, 2016.

\bibitem[Karpathy(2021)]{karpathy2021deepvectorquantization}
Andrej Karpathy.
\newblock deep-vector-quantization, 2021.
\newblock URL \url{https://github.com/karpathy/deep-vector-quantization}.
\newblock Accessed: 2024-09-30.

\bibitem[Kingma \& Welling(2013)Kingma and Welling]{kingma2013auto}
Diederik~P Kingma and Max Welling.
\newblock Auto-encoding variational bayes.
\newblock \emph{arXiv preprint arXiv:1312.6114}, 2013.

\bibitem[Krause et~al.(2010)Krause, Perona, and Gomes]{krause2010discriminative}
Andreas Krause, Pietro Perona, and Ryan Gomes.
\newblock Discriminative clustering by regularized information maximization.
\newblock \emph{Advances in neural information processing systems}, 23, 2010.

\bibitem[{\L}a{\'n}cucki et~al.(2020){\L}a{\'n}cucki, Chorowski, Sanchez, Marxer, Chen, Dolfing, Khurana, Alum{\"a}e, and Laurent]{lancucki2020robust}
Adrian {\L}a{\'n}cucki, Jan Chorowski, Guillaume Sanchez, Ricard Marxer, Nanxin Chen, Hans~JGA Dolfing, Sameer Khurana, Tanel Alum{\"a}e, and Antoine Laurent.
\newblock Robust training of vector quantized bottleneck models.
\newblock In \emph{2020 International Joint Conference on Neural Networks (IJCNN)}, pp.\  1--7. IEEE, 2020.

\bibitem[Liu et~al.(2021)Liu, Lamb, Kawaguchi, ALIAS PARTH~GOYAL, Sun, Mozer, and Bengio]{liu2021discrete}
Dianbo Liu, Alex~M Lamb, Kenji Kawaguchi, Anirudh~Goyal ALIAS PARTH~GOYAL, Chen Sun, Michael~C Mozer, and Yoshua Bengio.
\newblock Discrete-valued neural communication.
\newblock \emph{Advances in Neural Information Processing Systems}, 34:\penalty0 2109--2121, 2021.

\bibitem[Liu et~al.(2023)Liu, Liu, Li, Yu, Guo, Liu, and Wang]{liu2023cloud}
Yue Liu, Zitu Liu, Shuang Li, Zhenyao Yu, Yike Guo, Qun Liu, and Guoyin Wang.
\newblock Cloud-vae: Variational autoencoder with concepts embedded.
\newblock \emph{Pattern Recognition}, 140:\penalty0 109530, 2023.

\bibitem[Loshchilov(2017)]{loshchilov2017decoupled}
I~Loshchilov.
\newblock Decoupled weight decay regularization.
\newblock \emph{arXiv preprint arXiv:1711.05101}, 2017.

\bibitem[Maddison et~al.(2017)Maddison, Mnih, and Teh]{maddison2017the}
Chris~J. Maddison, Andriy Mnih, and Yee~Whye Teh.
\newblock The concrete distribution: A continuous relaxation of discrete random variables.
\newblock In \emph{International Conference on Learning Representations}, 2017.
\newblock URL \url{https://openreview.net/forum?id=S1jE5L5gl}.

\bibitem[Makhzani et~al.(2015)Makhzani, Shlens, Jaitly, Goodfellow, and Frey]{makhzani2015adversarial}
Alireza Makhzani, Jonathon Shlens, Navdeep Jaitly, Ian Goodfellow, and Brendan Frey.
\newblock Adversarial autoencoders.
\newblock \emph{arXiv preprint arXiv:1511.05644}, 2015.

\bibitem[Nalisnick et~al.(2016)Nalisnick, Hertel, and Smyth]{nalisnick2016approximate}
Eric Nalisnick, Lars Hertel, and Padhraic Smyth.
\newblock Approximate inference for deep latent gaussian mixtures.
\newblock In \emph{NIPS Workshop on Bayesian Deep Learning}, volume~2, pp.\  131, 2016.

\bibitem[Razavi et~al.(2019)Razavi, Van~den Oord, and Vinyals]{razavi2019generating}
Ali Razavi, Aaron Van~den Oord, and Oriol Vinyals.
\newblock Generating diverse high-fidelity images with vq-vae-2.
\newblock \emph{Advances in neural information processing systems}, 32, 2019.

\bibitem[Roy et~al.(2018)Roy, Vaswani, Neelakantan, and Parmar]{roy2018theory}
Aurko Roy, Ashish Vaswani, Arvind Neelakantan, and Niki Parmar.
\newblock Theory and experiments on vector quantized autoencoders.
\newblock \emph{arXiv preprint arXiv:1805.11063}, 2018.

\bibitem[Shu \& Nakayama(2017)Shu and Nakayama]{shu2017compressing}
Raphael Shu and Hideki Nakayama.
\newblock Compressing word embeddings via deep compositional code learning.
\newblock \emph{arXiv preprint arXiv:1711.01068}, 2017.

\bibitem[S{\o}nderby et~al.(2017)S{\o}nderby, Poole, and Mnih]{sonderby2017continuous}
Casper~Kaae S{\o}nderby, Ben Poole, and Andriy Mnih.
\newblock Continuous relaxation training of discrete latent variable image models.
\newblock In \emph{Beysian DeepLearning workshop, NIPS}, volume 201, 2017.

\bibitem[Takida et~al.(2022)Takida, Shibuya, Liao, Lai, Ohmura, Uesaka, Murata, Takahashi, Kumakura, and Mitsufuji]{takida2022sq}
Yuhta Takida, Takashi Shibuya, WeiHsiang Liao, Chieh-Hsin Lai, Junki Ohmura, Toshimitsu Uesaka, Naoki Murata, Shusuke Takahashi, Toshiyuki Kumakura, and Yuki Mitsufuji.
\newblock Sq-vae: Variational bayes on discrete representation with self-annealed stochastic quantization.
\newblock \emph{arXiv preprint arXiv:2205.07547}, 2022.

\bibitem[Tian et~al.(2024)Tian, Jiang, Yuan, Peng, and Wang]{tian2024visual}
Keyu Tian, Yi~Jiang, Zehuan Yuan, Bingyue Peng, and Liwei Wang.
\newblock Visual autoregressive modeling: Scalable image generation via next-scale prediction.
\newblock \emph{arXiv preprint arXiv:2404.02905}, 2024.

\bibitem[Tomczak \& Welling(2018)Tomczak and Welling]{tomczak2018vae}
Jakub Tomczak and Max Welling.
\newblock Vae with a vampprior.
\newblock In \emph{International conference on artificial intelligence and statistics}, pp.\  1214--1223. PMLR, 2018.

\bibitem[Van Den~Oord et~al.(2017)Van Den~Oord, Vinyals, et~al.]{van2017neural}
Aaron Van Den~Oord, Oriol Vinyals, et~al.
\newblock Neural discrete representation learning.
\newblock \emph{Advances in neural information processing systems}, 30, 2017.

\bibitem[Williams(1992)]{williams1992simple}
Ronald~J Williams.
\newblock Simple statistical gradient-following algorithms for connectionist reinforcement learning.
\newblock \emph{Machine learning}, 8:\penalty0 229--256, 1992.

\bibitem[Williams et~al.(2020)Williams, Ringer, Ash, MacLeod, Dougherty, and Hughes]{williams2020hierarchical}
Will Williams, Sam Ringer, Tom Ash, David MacLeod, Jamie Dougherty, and John Hughes.
\newblock Hierarchical quantized autoencoders.
\newblock \emph{Advances in Neural Information Processing Systems}, 33:\penalty0 4524--4535, 2020.

\bibitem[Wu et~al.(2024)Wu, Yan, and Liu]{wu2024vqsynery}
Jiawei Wu, Mingyuan Yan, and Dianbo Liu.
\newblock Vqsynery: Robust drug synergy prediction with vector quantization mechanism.
\newblock \emph{arXiv preprint arXiv:2403.03089}, 2024.

\bibitem[Yang et~al.(2021)Yang, Liu, Chen, Shen, Hao, and Wang]{yang2021causalvae}
Mengyue Yang, Furui Liu, Zhitang Chen, Xinwei Shen, Jianye Hao, and Jun Wang.
\newblock Causalvae: Disentangled representation learning via neural structural causal models.
\newblock In \emph{Proceedings of the IEEE/CVF conference on computer vision and pattern recognition}, pp.\  9593--9602, 2021.

\bibitem[Yu et~al.(2021)Yu, Li, Koh, Zhang, Pang, Qin, Ku, Xu, Baldridge, and Wu]{yu2021vector}
Jiahui Yu, Xin Li, Jing~Yu Koh, Han Zhang, Ruoming Pang, James Qin, Alexander Ku, Yuanzhong Xu, Jason Baldridge, and Yonghui Wu.
\newblock Vector-quantized image modeling with improved vqgan.
\newblock \emph{arXiv preprint arXiv:2110.04627}, 2021.

\bibitem[Yu et~al.(2023)Yu, Lezama, Gundavarapu, Versari, Sohn, Minnen, Cheng, Gupta, Gu, Hauptmann, et~al.]{yu2023language}
Lijun Yu, Jos{\'e} Lezama, Nitesh~B Gundavarapu, Luca Versari, Kihyuk Sohn, David Minnen, Yong Cheng, Agrim Gupta, Xiuye Gu, Alexander~G Hauptmann, et~al.
\newblock Language model beats diffusion--tokenizer is key to visual generation.
\newblock \emph{arXiv preprint arXiv:2310.05737}, 2023.

\bibitem[Zeghidour et~al.(2021)Zeghidour, Luebs, Omran, Skoglund, and Tagliasacchi]{zeghidour2021soundstream}
Neil Zeghidour, Alejandro Luebs, Ahmed Omran, Jan Skoglund, and Marco Tagliasacchi.
\newblock Soundstream: An end-to-end neural audio codec.
\newblock \emph{IEEE/ACM Transactions on Audio, Speech, and Language Processing}, 30:\penalty0 495--507, 2021.

\bibitem[Zhang et~al.(2024)Zhang, Bian, Chen, and Yao]{zhang2024unimot}
Juzheng Zhang, Yatao Bian, Yongqiang Chen, and Quanming Yao.
\newblock Unimot: Unified molecule-text language model with discrete token representation.
\newblock \emph{arXiv preprint arXiv:2408.00863}, 2024.

\bibitem[Zhang et~al.(2022)Zhang, Xiao, Paige, and Barber]{zhang2022improving}
Mingtian Zhang, Tim~Z Xiao, Brooks Paige, and David Barber.
\newblock Improving vae-based representation learning.
\newblock \emph{arXiv preprint arXiv:2205.14539}, 2022.

\bibitem[Zhao et~al.(2017)Zhao, Song, and Ermon]{zhao2017infovae}
Shengjia Zhao, Jiaming Song, and Stefano Ermon.
\newblock Infovae: Information maximizing variational autoencoders.
\newblock \emph{arXiv preprint arXiv:1706.02262}, 2017.

\end{thebibliography}
